\pdfoutput=1
\documentclass[11pt]{article}
\usepackage[preprint]{acl}
\usepackage{times}
\usepackage{latexsym}
\usepackage[T1]{fontenc}
\usepackage[utf8]{inputenc}
\usepackage{microtype}
\usepackage{inconsolata}
\usepackage{graphicx}
\usepackage{amsmath}
\usepackage{amsfonts}
\usepackage{booktabs}
\usepackage{multirow}
\usepackage{xcolor}
\usepackage{comment}
\usepackage{enumitem}
\usepackage{pifont}
\usepackage{float}
\setlist[enumerate]{noitemsep, topsep=0pt}

\usepackage{fancyhdr}
\title{Beyond Pixels: Introspective and Interactive Grounding for \\ Visualization Agents\thanks{Code and data available at \url{https://github.com/HexSys-lab/Deep-Data} and \url{https://github.com/HexSys-lab/iPlotBench}}}

\author{Yiyang Lu\textsuperscript{1} \quad Woong Shin\textsuperscript{2} \quad Ahmad Maroof Karimi\textsuperscript{2} \\
\textbf{Feiyi Wang\textsuperscript{2} \quad Jie Ren\textsuperscript{1,*} \quad Evgenia Smirni\textsuperscript{1,*}} \\
\textsuperscript{1}William \& Mary \quad \textsuperscript{2}Oak Ridge National Laboratory\\
\texttt{\{ylu21, exsmir\, jren03\}@wm.edu} \quad \texttt{\{shinw, karimiahmad, fwang2\}@ornl.gov}}

\begin{document}
\maketitle
\thispagestyle{empty}
\begingroup
\renewcommand\thefootnote{*}
\footnotetext{These authors jointly supervised this work.}
\endgroup

\begin{abstract}

Vision-Language Models (VLMs) frequently misread values, hallucinate details, and confuse overlapping elements in charts. Current approaches rely solely on pixel interpretation, creating a Pixel-Only Bottleneck: agents treat interactive charts as static images, losing access to the structured specification that encodes exact values. We introduce Introspective and Interactive Visual Grounding (IVG), a framework that combines (1) spec-grounded introspection, which queries the underlying specification for deterministic evidence, with (2) view-grounded interaction, which manipulates the view to resolve visual ambiguity. To enable evaluation without VLM bias, we present iPlotBench, a benchmark of 500 interactive Plotly figures with 6,706 binary questions and ground-truth specifications. Experiments show that introspection improves data reconstruction fidelity, while the combination with interaction achieves the highest QA accuracy (0.81), with +6.7\% gains on overlapping geometries. We further demonstrate IVG in deployed agents that explore data autonomously and collaborate with human users in real time.

\end{abstract}

\section{Introduction}

Large language models are increasingly trusted to answer complex queries, analyze technical documents, and assist in high-stakes decision-making. Yet this trust is fragile. A persistent limitation is hallucination: generating fluent, confident text that fabricates facts, misremembers details, or contradicts sources. As we move toward multimodal systems, Vision-Language Models inherit and compound this problem. While VLMs extend the capabilities of LLMs to the visual domain, they introduce new forms of error. They exhibit perceptual hallucinations: misreading numbers, confusing visually similar elements, or reporting details not present in the image~\cite{kaul2024throne, jiang2024hallucination, rohrbach2018object, li2023evaluating, leng2024mitigating}. In domains like data analysis and chart interpretation, these errors are not cosmetic---they can lead to fundamentally flawed conclusions. Unlike text, where a hallucination might be a subtle twist of phrase, a VLM misreading a trend line or data point breaks the chain of reasoning entirely.

For photographs, resolving such errors is difficult---there is no structured ground truth to verify against. But charts are different. A chart is rendered from a specification that encodes exact values, bindings, and relationships. Whether agents generate charts, answer questions about them, or refine visualizations based on feedback, they could consult this specification directly. Yet today's agents never do. They interpret charts through pixels alone---a static, lossy representation where values become approximate and overlapping elements become ambiguous~\cite{yangchartmimic, zhao2025chartcoder, seo2025automatedvisualizationcodesynthesis, hong-etal-2025-data, li2025metal}. The specification contains exact data, axis ranges, and trace definitions, yet it goes unused. The view is fixed---agents cannot zoom, toggle, or select to resolve ambiguity. We call this the \textbf{Pixel-Only Bottleneck}: reasoning is limited to static pixels, with no access to the underlying specification and no ability to manipulate the view.

\begin{figure*}[htbp]
\centering
\includegraphics[width=\textwidth]{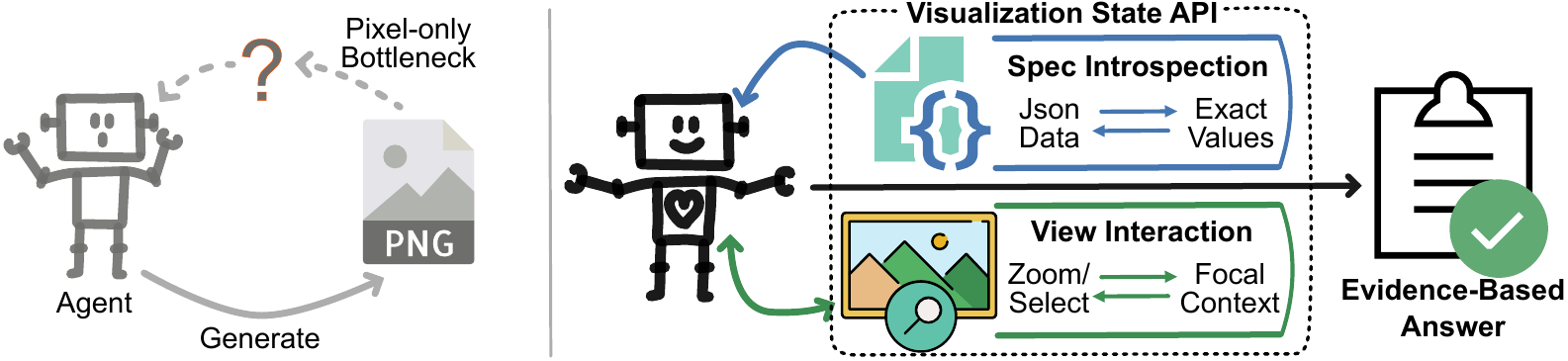}
\caption{Overview of IVG. Rather than reasoning by interpreting rendered pixels with a VLM (left), the agent accesses the visualization state to (i) perform \textbf{spec-grounded introspection} over the chart specification for deterministic evidence and (ii) perform \textbf{view-grounded interaction} (e.g., zoom, toggle, select) to obtain focal context in visually ambiguous regions. This transforms agents from passive observers into active owners of their visualizations.}
\vspace{-5pt}
\label{fig:ivg_overview}
\end{figure*}

Humans routinely resolve such ambiguity by interacting with charts: zooming, toggling traces, and inspecting exact values. Why not give agents the same capabilities? Visualization libraries such as Plotly, Vega-Lite, and ECharts render charts from structured specifications that separate data values from the rendered view. This architecture makes it possible for agents to operate on the visualization's source of truth rather than guessing from pixels. We introduce \textbf{Introspective and Interactive Visual Grounding (IVG)}, a framework that equips agents with two complementary mechanisms:

\begin{itemize}[leftmargin=*, topsep=2pt, itemsep=1pt]
    \item \textbf{Spec-grounded introspection} provides direct access to the chart's underlying specification, allowing agents to verify exact values and attributes.
    \item \textbf{View-grounded interaction} provides tools for manipulating the view (zoom, pan, toggle, select), allowing agents to resolve ambiguity in dense or overlapping regions. Each interaction produces \textit{focal context}---axis ranges, trace indices, or selected points---that tells the agent where to look in the specification.
\end{itemize}

Together, these mechanisms enable grounded reasoning at two levels. 
Between agent and data, IVG transforms agents from passive observers into active explorers of their own visualizations. Between human and agent, the same visualization state serves as a shared communication channel: a user's interactions provide context that the agent can interpret and respond to, without the user needing to verbally describe what they see (Figure \ref{fig:ivg_overview}). IVG is implemented as a Visualization State API using Model Context Protocol (MCP) tools, allowing any compatible agent to access these capabilities.

Evaluating such capabilities is challenging. Existing benchmarks like ChartQA~\cite{masry2022chartqa} and ChartMimic~\cite{yangchartmimic} treat charts as static images and rely on VLM-based evaluation, but VLMs make the same perceptual errors we want to avoid. 
We introduce \textbf{iPlotBench}, a benchmark of 500 interactive Plotly figures with 6,706 binary questions and ground-truth specifications, designed for full deterministic evaluation. Our metric, \textbf{Semantic Structural Similarity}, evaluates chart recreation by directly comparing agent-produced and ground-truth specifications.

In summary, our contributions are as follows:
\begin{enumerate}
    \item \textbf{IVG Toolkit:} A set of MCP tools that let agents query chart specifications and manipulate views, enabling deterministic reasoning over visualizations.
    \item \textbf{iPlotBench:} A benchmark of 500 interactive Plotly figures with 6,706 binary questions, ground-truth specifications, and Semantic Structural Similarity for deterministic evaluation.
    \item Our experiments on Claude Haiku 4.5 and Qwen demonstrate that spec-grounded introspection improves data reconstruction fidelity ($S_{Data}$: 0.88$\rightarrow$0.90), while view-grounded interaction yields the largest QA accuracy gains (+6.7\%) on questions involving complex overlapping geometries. The combination achieves the highest overall QA accuracy.
    \item Beyond controlled evaluation, we deploy IVG in three agent workflows: real-time collaboration, autonomous exploration, and ML solution search. These deployments demonstrate that spec-grounded introspection and view-grounded interaction serve as general-purpose capabilities across diverse analytical tasks. 
\end{enumerate}

\section{Related Work}

\paragraph{Chart Understanding and QA.}
Benchmarks like FigureQA \citep{figureqa} and DVQA \citep{kafle2018dvqa} establish the task of Visual Question Answering (VQA) on synthetic plots. Subsequent datasets like ChartQA \citep{masry2022chartqa} and PlotQA \citep{methani2020plotqa} extend this to real-world charts requiring complex reasoning. However, these approaches treat charts as static pixel arrays, forcing models to rely on visual approximations. Our work moves beyond static pixel interpretation by enabling agents to query chart specifications and interact with views directly.

\paragraph{Chart-to-Code and Recreation.}
Existing work focuses on agents that generate code to render charts. ChartMimic \citep{yangchartmimic}, Plot2code \citep{wu2025plot2code}, Chart2Code53 \citep{niu2025chart2code53} and ChartCoder \citep{zhao2025chartcoder} evaluate agents on their ability to visually reproduce reference charts. However, these benchmarks rely on VLM-based evaluation (e.g., GPT-4V) or pixel-level metrics, which cannot verify whether the underlying data values are correct. We introduce \textit{Semantic Structural Similarity}, a metric that deterministically evaluates chart reconstruction by directly comparing the agent-produced Plotly JSON against the released ground-truth specification.

\begin{figure*}[htbp]
    \centering
    \includegraphics[width=\textwidth]{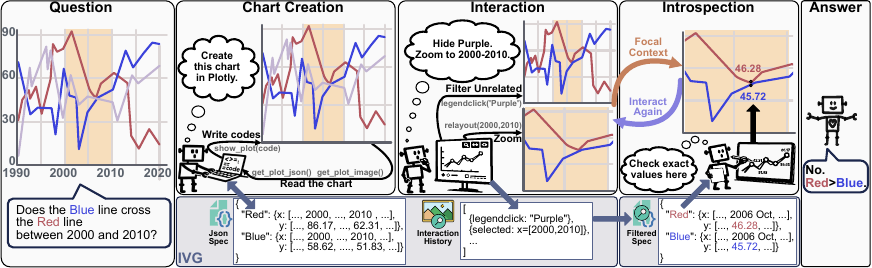}
    \caption{IVG workflow on a concrete example. Given a question, the agent recreates the chart (Chart Creation), gaining access to its specification and interaction tools. It then alternates between Interaction (toggling traces, zooming to generate focal context) and Introspection (querying the specification for exact values in the focused region) until sufficient evidence is gathered to produce a grounded answer.}
    \vspace{-5pt}
    \label{fig:ivg_workflow}
\end{figure*}

\paragraph{Tool-Integrated Coding Agents.}
Coding agents \citep{zhang2024codeagent, yang2024sweagent, guorepoaudit, wang2024opendevin, zhang2024autocoderover, xia2024agentless, schick2023toolformer, yao2022react} demonstrate the power of structured state access for code-related tasks, using interfaces such as terminals, file systems, and linters. However, these agents lack structured access to their visual outputs: a generated chart is interpreted back through pixels, discarding the specification that produced it. IVG extends structured state access to visualization. 

\paragraph{Interactive Visual Analytics.}
The principle that analysts should interact with visualizations rather than passively observe them is well established in the visualization community. Systems such as Voyager~\cite{Voyager1, Voyager2} and Draco~\cite{draco} leverage visualization grammars to generate chart designs from specifications. Conversely, IVG helps agents reason about existing charts by reading specifications back and returning structured state that agents can compose with specification queries. 
\section{IVG Design}

IVG enables agents to shift reasoning from probabilistic pixel inspection to deterministic, structured access. It builds on a key property of modern visualization libraries: libraries such as Plotly, Vega-Lite, and ECharts render charts from structured specifications that encode exact data values and visual attributes~\cite{satyanarayan2016vega, zhu2013data, wickham2011ggplot2, wilkinson2011grammar}. As illustrated in Figure~\ref{fig:ivg_workflow}, IVG combines \textit{spec-grounded introspection} (reading the underlying specification) with \textit{view-grounded interaction} (changing the rendered view to elicit structured state) into an iterative workflow for resolving visual ambiguity.

Spec-grounded introspection gives agents direct access to the JSON specification underlying a chart. This specification encodes the complete chart state: data arrays, bindings, and visual attributes. Rather than estimating values from pixels, agents query the specification directly. To verify whether a bar represents the maximum, the agent extracts the data array and computes the answer symbolically. To confirm a legend entry, the agent reads the corresponding field rather than interpreting rendered text. This directness matters because pixel-based perception is inherently approximate. Two bars of similar height may be indistinguishable visually but differ meaningfully in value; the specification reveals the exact numbers. By grounding verification in structured data, agents avoid the estimation errors and hallucinations that arise from pixel interpretation.

While spec-grounded introspection provides exact values, it requires knowing which part of the specification to examine. A chart may contain hundreds of points across multiple traces; retrieving and reasoning over the full specification is often unnecessary. View-grounded interaction addresses this challenge by generating \textbf{focal context}---explicit view-state parameters that constrain the search space. For example, zooming or panning yields precise \textbf{axis ranges} (e.g., \texttt{layout.xaxis.range}); toggling a trace identifies a \textbf{trace/curve index} (e.g., \texttt{curve\_number}); selecting a region yields a \textbf{data-space bounding box} and the points it contains. These interaction outputs act as \textbf{search keys}: instead of parsing the entire specification, the agent can filter the JSON returned by introspection to the subset relevant to its current focus. This transforms interaction history into a structured query mechanism that bridges visual attention and specification data.

These two mechanisms work together in a unified workflow, as shown in figure~\ref{fig:ivg_workflow}. If the relevant data can be directly located, introspection suffices. Otherwise, the agent first uses interaction to obtain focal context, then queries the specification within that scope. For example, to determine whether two lines intersect, an agent zooms into the region where they appear close, then queries for exact coordinates. The workflow is iterative: the agent refines its focus until sufficient evidence emerges.

The Visualization State API is implemented as a set of MCP tools, summarized in Table~\ref{tab:tools}. The interaction tools cover the three fundamental view-manipulation primitives identified by existing work~\cite{6634168} : navigate (relayout), select (selected), and filter (legendclick). These tools operate at the figure level and propagate across subplots with shared axes, supporting multi-view analysis. To ensure valid evaluation, agents can only access specifications and interaction state for figures they create (via returned \texttt{plot\_id}s), and they do not receive ground-truth datasets or reference solutions through the tool interface.

\begin{table}[t]
\centering
\scriptsize
\resizebox{\columnwidth}{!}{%
\begin{tabular}{llll}
\toprule
\textbf{Category} & \textbf{Tool} & \textbf{Args} & \textbf{Function} \\
\midrule
\multirow{2}{*}{Base}
  & show\_plot & plotly\_codes & Creates figure; returns plot\_id \\
  & get\_plot\_image & plot\_id & Returns current view as PNG \\
\midrule
Introspection
  & get\_plot\_json & plot\_id & Returns full Plotly specification \\
\midrule
\multirow{4}{*}{Interaction}
  & relayout & plot\_id, x/y\_range & Zoom/Pan axis bounds \\
  & legendclick & plot\_id, curve\_number & Toggle trace visibility \\
  & selected & plot\_id, x/y\_range & Returns points in region \\
  & query\_interactions & plot\_id & Returns interaction history \\
\bottomrule
\end{tabular}%
}
\caption{Visualization State API grouped by category.}
\vspace{-10pt}
\label{tab:tools}
\end{table}

\section{iPlotBench}

Existing visualization benchmarks, such as ChartMimic \citep{yangchartmimic} and FigureQA \citep{figureqa}, treat charts as static images. This makes it difficult to evaluate agents that (1) use view interaction (e.g., zooming into dense regions or toggling traces) and (2) require a rendered chart specification for deterministic, structure-level evaluation, rather than pixel- or VLM-based scoring.

To bridge this gap, we introduce iPlotBench, a benchmark for evaluating \textbf{spec-grounded introspection} and \textbf{view-grounded interaction} in visualization agents. iPlotBench procedurally generates interactive Plotly figures and releases their ground-truth specifications, enabling deterministic evaluation of chart recreation and visual reasoning.

\subsection{Dataset Construction}
The benchmark has 500 figures balanced across five chart types: Line, Dot-Line, Vertical Bar, Horizontal Bar, and Pie. We adapt the procedural generation pipeline from FigureQA \citep{figureqa} and convert each instance into an interactive Plotly figure. All figures support standard interactions (zoom/pan, legend toggling, selection), enabling agents to actively probe ambiguous regions.

Each figure is paired with a rendered PNG input and its ground-truth Plotly JSON specification. The dataset contains 6,706 binary questions across 15 templates, with a rigorous 50/50 Yes/No balance per figure to prevent class-imbalance shortcuts (Table~\ref{tab:dataset_stats}).

\begin{table}[t]
\centering
\small
\begin{tabular}{lrrr}
\toprule
Type & \#Figures & \#Questions & Q/Fig \\
\midrule
Line & 100 & 1{,}504 & 15.0 \\
Dot-Line & 100 & 1{,}690 & 16.9 \\
Vertical Bar & 100 & 1{,}142 & 11.4 \\
Horizontal Bar & 100 & 1{,}187 & 11.9 \\
Pie & 100 & 1{,}183 & 11.8 \\
\midrule
Total & 500 & 6{,}706 & 13.4 \\
\bottomrule
\end{tabular}
\caption{Dataset statistics. The benchmark features a balanced distribution of questions across five chart types, with an average of 13.4 questions per figure.}
\label{tab:dataset_stats}
\end{table}

The 15 question templates span three types: \textit{Aggregation} (min/max/median, e.g., ``Is Blue Violet the minimum?''), \textit{Comparison} (value ordering, e.g., ``Is Blue less than Red?''), and \textit{Topology} (intersection/smoothness, e.g., ``Does Blue intersect Red?'').

\subsection{Tasks}
\label{sec:tasks}
We define two evaluation tasks that target complementary capabilities. In our evaluation, the tasks are sequential within the same agent session: the agent first recreates the chart (Task~1), then answers questions based on the recreated chart (Task~2).

\paragraph{Task 1: Chart Recreation.} Given a static reference image, the agent must generate the underlying Plotly JSON specification that faithfully recreates the chart. Unlike previous ``chart-to-code'' tasks that allow arbitrary styling, this task requires precise structural grounding: mapping visual elements (e.g., bar heights, colors, labels) to exact attributes in the JSON schema. Success in this task demonstrates the agent's ability to accurately reconstruct the structured specification from a visual reference.

\paragraph{Semantic structural similarity}
For Task 1, we propose Semantic Structural Similarity, which evaluates semantic correctness of agent-generated Plotly figures via direct JSON inspection. Unlike recent benchmarks such as ChartMimic which rely on VLM-based scoring (e.g., GPT-4o) to assess data trends and can be sensitive to VLM hallucinations, our metrics provide deterministic, symbolic verification. Exact JSON matching is impractical due to trace reordering and variable sampling density (e.g., different discretizations of the same curve). We therefore align predicted traces ($P$) to ground-truth traces ($T$) using Hungarian matching on data geometry: we compute pairwise Chamfer distances and solve the linear assignment to obtain a set of matched trace pairs $Pairs$. All trace-based metrics use the shared denominator $\max(|T|, |P|)$ to penalize extra/missing traces; unmatched traces contribute zero.

We intentionally omit separate layout and clarity metrics: in our single-plot figures, layout is largely captured by role-aware text, while clarity is subjective and is typically reflected as data/style errors. 

\paragraph{Chart Type ($S_{Type}$)} We verify the trace type (e.g., \textit{scatter}, \textit{bar}) for each matched pair, scoring 1 for a match and 0 otherwise, normalized by $\max(|T|, |P|)$ .

\paragraph{Data Accuracy ($S_{Data}$)} We treat each trace as a point cloud and compute Chamfer distance between matched pairs, robust to sampling differences. Within Chamfer, we use Euclidean distance on range-normalized numerical dimensions and a Jaccard-based penalty for categorical/text dimensions (e.g., bar labels):
\begin{equation}
    S_{Data} = \sum_{(t,p) \in Pairs} \frac{e^{-\lambda D_{Chamfer}(t,p)}}{\max(|T|, |P|)} 
\end{equation}
We set $\lambda=5$ to balance sensitivity: at $\lambda=1$, even worst-case reconstructions score 0.37, whereas at $\lambda=10$, minor noise is penalized too heavily. A sensitivity analysis over $\lambda$ confirms that configuration rankings remain consistent across all values (Section~\ref{sec:lambda_sensitivity}).

\paragraph{Text Correctness ($S_{Text}$)} We use a role-aware text metric to avoid ``right text, wrong place'' errors. Text is bucketed into semantic roles $\mathcal{R}$ (Title, Axis, Legend, Data/Annotations), and compared with a fuzzy Jaccard similarity:
\begin{equation}
    S_{Text} = \sum_{r \in \mathcal{R}} w_r \cdot \text{Sim}(Text_T^r, Text_P^r)
\end{equation}

\paragraph{Visual Style ($S_{Style}$)} We average style similarity over matched traces and properties $\mathcal{P}=\{\textit{color}, \textit{mode}, \textit{symbol}, \textit{size}, \textit{dash}, \textit{width}\}$:
\begin{equation}
    S_{Style} = \frac{\sum_{(t,p) \in Pairs} \sum_{prop \in \mathcal{P}} S_{prop}(t, p)}{\max(|T|, |P|)\,|\mathcal{P}|}
\end{equation}
Here $S_{prop}(t,p)$ compares the values of property $prop$ in traces $t$ and $p$. For color, we compare in CIELAB space (and use EMD for color arrays); categorical attributes use exact match and numerical attributes use normalized error, comparing against Plotly defaults when an attribute is omitted.

\paragraph{Task 2: Visual QA.}  
Given a chart and a binary question, the agent must answer yes or no; we report standard accuracy. Many questions are designed to be visually ambiguous in a static view (e.g., detecting intersections in overlapping line charts or identifying the minimum in a dense bar chart). We use binary questions since they enable fully deterministic evaluation without VLM-based or human judgment. Open-ended reasoning with IVG is demonstrated in Section~\ref{sec:deepplot}.

\section{Evaluation}

\subsection{Evaluation Setup}
 
We compare four configurations: \textit{Vision} (no tools), \textit{+Inter} (interaction-only tools), \textit{+Intro} (introspection-only via \textit{get\_plot\_json}), and \textit{Full} (both introspection and interaction tools). 

Agents follow the two-task protocol in Section~\ref{sec:tasks}. Tools are available but tool use is optional; all configurations share the same prompt template and tool schemas (Appendix~\ref{sec:appendix_controller}).

\textbf{Models. }
Our main ablation uses Claude Haiku 4.5. We also evaluate the Qwen-VL family (Table~\ref{tab:qwen_models}) for model scaling analysis. Because provider runtimes may include undisclosed system prompts, we emphasize \textit{within-model} deltas (e.g., Vision vs Full) for causal claims.

\begin{table}[h]
\centering
\scriptsize
\begin{tabular}{lll}
\toprule
Model & Architecture & Active Params \\
\midrule
Qwen3-VL-30B-A3B & MoE & 3B \\
Qwen3-VL-32B & Dense & 32B \\
Qwen3-VL-235B-A22B & MoE & 22B \\
Qwen-VL-Max & multimodal VLM  & $>$72B (est.) \\
\bottomrule
\end{tabular}
\caption{Qwen-VL models evaluated.
}
\vspace{-5pt}
\label{tab:qwen_models}
\end{table}

\textbf{Results summary. } We evaluate four agent configurations on iPlotBench. Within each episode, the agent first recreates the chart from the static reference image (Task~1), then answers binary questions by reasoning over its recreated interactive figure (Task~2). Overall, spec-grounded introspection primarily improves chart reconstruction, while combining introspection with interaction yields the strongest QA performance.

\subsection{Task 1: Chart Recreation}

\noindent As shown in Table~\ref{tab:task1}, equipping agents with introspection (\textit{+Intro}) drives the most significant gains in semantic reconstruction, achieving the highest scores in trace typing ($S_{Type}$), data fidelity ($S_{Data}$), and style ($S_{Style}$).
While interaction (\textit{+Inter}) slightly outperforms in text extraction ($S_{Text}$), likely by exposing occluded labels, it provides limited structural benefit.
Notably, the \textit{Full} agent does not surpass \textit{+Intro}: adding interaction expands the action space, and unnecessary view changes can distract from spec-level fixes during reconstruction.
\begin{table}[h]
\centering
\tiny
\resizebox{\columnwidth}{!}{
\begin{tabular}{lcccc}
\toprule
Metric & Vision & +Inter & +Intro & Full \\
\midrule
$S_{Type}$ & 0.9542 & 0.9742 & \textbf{0.9807} & 0.9744 \\
$S_{Data}$ & 0.8847 & 0.8935 & \textbf{0.9016} & 0.8907 \\
$S_{Text}$ & 0.9755 & \textbf{0.9903} & 0.9847 & 0.9877 \\
$S_{Style}$ & 0.8699 & 0.8708 & \textbf{0.8749} & 0.8679 \\
\bottomrule
\end{tabular}%
}
\caption{Semantic Structural Similarity for Task~1.}
\vspace{-10pt}
\label{tab:task1}
\end{table}
\subsection{Task 2: Visual QA}

\noindent Table~\ref{tab:task2} highlights the synergy of introspection and interaction: the \textit{Full} agent achieves the highest overall accuracy (0.8062).
The benefits of interaction are most pronounced for Line and Dot-Line charts, where zooming and selective inspection help resolve overlaps. For Bar and Pie charts, \textit{+Intro} achieves the best accuracy, suggesting that direct spec access is sufficient when the geometry is less ambiguous.
\begin{table}[h]
\centering
\small
\resizebox{\columnwidth}{!}{
\begin{tabular}{lcccc}
\toprule
Figure Type & Vision & +Inter & +Intro & Full \\
\midrule
Line & 0.7012 & 0.7030 & 0.7085 & \textbf{0.7292} \\
Dot-Line & 0.7155 & 0.7178 & 0.7353 & \textbf{0.7522} \\
Vertical Bar & 0.7993 & 0.8261 & \textbf{0.8529} & 0.8434 \\
Horizontal Bar & 0.8354 & 0.8388 & \textbf{0.8623} & 0.8463 \\
Pie & 0.8986 & 0.9019 & \textbf{0.9197} & 0.9146 \\
\midrule
OVERALL & 0.7783 & 0.7850 & 0.8023 & \textbf{0.8062} \\
\bottomrule
\end{tabular}%
}
\caption{Question-level accuracy for Task~2.}
\vspace{-10pt}
\label{tab:task2}
\end{table}

\subsection{Deconfounding Reconstruction Quality}

Since Task~2 is performed on the chart recreated in Task~1, QA accuracy is bounded by reconstruction fidelity. To isolate reasoning, we condition on well-reconstructed figures ($S_{Data} \ge 0.9$). Table~\ref{tab:conditional_qa} reports per-figure accuracy, which averages accuracy within each figure before averaging across figures to avoid overweighting chart types with more questions. 

\begin{table}[h]
\centering
\small
\resizebox{\columnwidth}{!}{
\begin{tabular}{lcc}
\toprule
Config & Per-figure Acc & Cond. Acc ($S_{Data} \ge 0.9$) \\
\midrule
Vision & 0.7941 & 0.8321 \\
+Inter & 0.8013 & 0.8538 \\
+Intro & 0.8190 & 0.8606 \\
Full & \textbf{0.8199} & \textbf{0.8637} \\
\bottomrule
\end{tabular}%
}
\caption{Per-figure and conditional QA accuracy.}
\vspace{-5pt}
\label{tab:conditional_qa}
\end{table}

\noindent Table~\ref{tab:family} reveals complementary strengths: \textit{+Intro} excels at Comparison questions where direct spec access enables precise value retrieval, while \textit{Full} performs best on Topology questions where interaction helps resolve spatial relationships.

\begin{table}[h]
\centering
\small
\begin{tabular}{lccc}
\toprule
Config & Aggregation & Comparison & Topology \\
\midrule
Vision & 0.8354 & 0.8391 & 0.7378 \\
+Inter & 0.8591 & 0.8522 & 0.8043 \\
+Intro & 0.8846 & \textbf{0.8555} & 0.7567 \\
Full & \textbf{0.8864} & 0.8514 & \textbf{0.8052} \\
\bottomrule
\end{tabular}
\caption{QA accuracy on well-reconstructed figures. Aggregation: min/max/median; Comparison: less/greater; Topology: intersection, smoothness, AUC.}
\vspace{-5pt}
\label{tab:family}
\end{table}

\begin{table}[H]
\centering
\scriptsize
\begin{tabular}{llcccc}
\toprule
\multirow{2}{*}{Config} & \multirow{2}{*}{Chart Type} & \texttt{relayout} & \texttt{legend} & \texttt{selected} & Total \\
 & & (\%) & (\%) & (\%) & (calls) \\
\midrule
\multirow{5}{*}{+Inter}
 & Line           & 92 \% & 52 &  38 & 349 \\
 & Dot-Line       & 88 & 53 &  29 & 343 \\
 & Vertical Bar   & 51 &  3 &  12 &  82 \\
 & Horizontal Bar & 59 &  2 &  11 &  92 \\
 & Pie            &  1 & 31 &   5 &  62 \\
\midrule
\multirow{5}{*}{Full}
 & Line           & 32 & 29 &  19 & 112 \\
 & Dot-Line       & 25 & 25 &  12 &  90 \\
 & Vertical Bar   & 13 &  3 &   5 &  27 \\
 & Horizontal Bar & 15 &  1 &   5 &  25 \\
 & Pie            &  0 &  7 &   1 &  13 \\
\bottomrule
\end{tabular}
\caption{View-manipulation tool usage by chart type and configuration. Columns show the percentage of sessions invoking each tool; Total is the number of calls.}
\label{tab:tool_usage}
\end{table}
\subsection{IVG Usage Analysis}
\label{sec:usage_analysis}
To understand when and why each mechanism helps, Table~\ref{tab:tool_usage} reports interaction tool usage across agent sessions by chart type and configuration.

Agents use zoom/pan (relayout), trace toggling (legend), and region selection (selected) heavily on Line and Dot-Line charts, where overlapping traces create spatial ambiguity, but rarely on Bar and Pie charts, where values are visually distinct. When introspection is also available (Full), total interaction calls drop sharply, indicating that agents rely on direct spec access instead of view manipulation when both options are available. This pattern is consistent across Tasks 1 and 2, and explains why Full underperforms \textit{+Intro} on $S_{Data}$ (in Table~\ref{tab:task1}): unnecessary interactions consume the tool-call budget without providing information beyond what the specification already contains.

\subsection{$\lambda$ Sensitivity Analysis}
\label{sec:lambda_sensitivity}

We evaluate the sensitivity of $S_{Data}$ to the exponential penalty parameter $\lambda$ in Equation~2 by re-computing scores with $\lambda \in \{1, 3, 5, 7, 10\}$ across both model families.

Table~\ref{tab:lambda_haiku} reports results for Claude Haiku 4.5. The ranking of configurations is preserved across all $\lambda$ values: \textit{+Intro} consistently achieves the highest $S_{Data}$, followed by Full, +Inter, and Vision.

\begin{table}[h]
\centering
\small
\begin{tabular}{lccccc}
\toprule
Config & $\lambda\!=\!1$ & $\lambda\!=\!3$ & $\lambda\!=\!5$ & $\lambda\!=\!7$ & $\lambda\!=\!10$ \\
\midrule
Vision & 0.9603 & 0.9191 & 0.8847 & 0.8548 & 0.8158 \\
+Inter & 0.9719 & 0.9297 & 0.8935 & 0.8618 & 0.8203 \\
+Intro & \textbf{0.9742} & \textbf{0.9352} & \textbf{0.9016} & \textbf{0.8720} & \textbf{0.8327} \\
Full   & 0.9696 & 0.9268 & 0.8907 & 0.8592 & 0.8182 \\
\bottomrule
\end{tabular}
\caption{$S_{Data}$ across $\lambda$ values for Claude Haiku 4.5 configurations.}
\label{tab:lambda_haiku}
\end{table}

Table~\ref{tab:lambda_qwen} reports results for the Qwen-VL family. Full consistently improves over Vision for models with sufficient capacity (Qwen3-VL-235B-A22B, Qwen-VL-Max) and degrades for the smallest model, consistent with the tool capability threshold discussed in Section~\ref{sec:qwen_scaling}. Minor ranking variations occur at boundary cases (e.g., Qwen3-VL-32B at $\lambda=7$, margin of 0.004), but overall patterns hold across all $\lambda$ values.

\begin{table}[h]
\centering
\small
\resizebox{\columnwidth}{!}{
\begin{tabular}{llccccc}
\toprule
Model & Config & $\lambda\!=\!1$ & $\lambda\!=\!3$ & $\lambda\!=\!5$ & $\lambda\!=\!7$ & $\lambda\!=\!10$ \\
\midrule
\multirow{2}{*}{Qwen3-VL-30B-A3B}
 & Vision & 0.959 & 0.899 & 0.852 & 0.813 & 0.765 \\
 & Full   & 0.783 & 0.667 & 0.642 & 0.630 & 0.616 \\
\midrule
\multirow{2}{*}{Qwen3-VL-32B}
 & Vision & 0.939 & 0.907 & 0.881 & 0.859 & 0.831 \\
 & Full   & 0.959 & 0.916 & 0.883 & 0.855 & 0.821 \\
\midrule
\multirow{2}{*}{Qwen3-VL-235B-A22B}
 & Vision & 0.918 & 0.897 & 0.878 & 0.860 & 0.836 \\
 & Full   & \textbf{0.962} & \textbf{0.929} & \textbf{0.904} & \textbf{0.881} & \textbf{0.851} \\
\midrule
\multirow{2}{*}{Qwen-VL-Max}
 & Vision & 0.965 & 0.939 & 0.916 & 0.896 & 0.868 \\
 & Full   & \textbf{0.971} & \textbf{0.945} & \textbf{0.922} & \textbf{0.901} & \textbf{0.874} \\
\bottomrule
\end{tabular}
}
\caption{$S_{Data}$ across $\lambda$ values for Qwen-VL models (Vision vs.\ Full).}
\label{tab:lambda_qwen}
\end{table}

\subsection{Model Scaling Analysis}
\label{sec:qwen_scaling}

We examine how IVG benefits scale with model capability using the Qwen-VL family. Table~\ref{tab:qwen_main} presents Task~1 ($S_{Data}$) and Task~2 (QA accuracy) for Vision and Full configurations. We observe that IVG benefits scale with model capability. Models with larger active parameters (Qwen3-VL-235B-A22B, Qwen-VL-Max) show consistent improvements, while smaller models face challenges in effectively orchestrating tool use alongside their primary reasoning tasks.

\begin{table}[h]
\centering
\small
\resizebox{\columnwidth}{!}{
\begin{tabular}{lcccc}
\toprule
Model & $S_{Data}$ (V) & $S_{Data}$ (F) & QA (V) & QA (F) \\
\midrule
Qwen3-VL-30B-A3B & 0.852 & 0.642 & 0.948 & 0.870 \\
Qwen3-VL-32B & 0.881 & \textbf{0.883} & 0.956 & 0.933 \\
Qwen3-VL-235B-A22B & 0.878 & \textbf{0.904} & 0.951 & \textbf{0.960} \\
Qwen-VL-Max & 0.916 & \textbf{0.922} & 0.971 & \textbf{0.973} \\
\bottomrule
\end{tabular}
}
\caption{Task~1 and Task~2 results for Qwen-VL models. V=Vision (no tools), F=Full (IVG tools).
}
\label{tab:qwen_main}
\end{table}

\begin{figure*}[!htbp]
    \centering
    \includegraphics[width=0.9\textwidth]{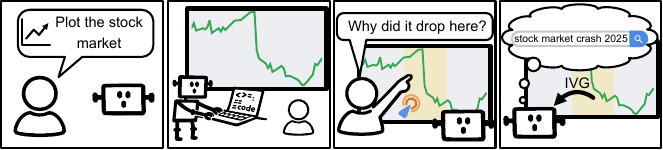}
    \caption{Real-time collaboration. The user points at a region of interest and asks a question; IVG captures this as focal context, enabling the agent to understand what the user is referring to.}
    \vspace{-5pt}
    \label{fig:interactive_workflow}
\end{figure*}

\subsection{Model Scaling: Conditional Analysis}
\label{sec:qwen_conditional}

To isolate verification capability from reconstruction quality, Table~\ref{tab:qwen_conditional} reports QA accuracy conditioned on successful reconstruction ($S_{Data} \geq 0.9$).

\begin{table}[h]
\centering
\small
\resizebox{\columnwidth}{!}{
\begin{tabular}{lccc}
\toprule
Model & Cond. (V) & Cond. (F) & $\Delta$ \\
\midrule
Qwen3-VL-30B-A3B & 0.959 & 0.900 & $-0.059$ \\
Qwen3-VL-32B & 0.973 & 0.964 & $-0.009$ \\
Qwen3-VL-235B-A22B & 0.961 & \textbf{0.966} & $+0.005$ \\
Qwen-VL-Max & 0.983 & \textbf{0.985} & $+0.002$ \\
\bottomrule
\end{tabular}
}
\caption{Conditional QA accuracy ($S_{Data} \ge 0.9$) for Qwen-VL models.}
\label{tab:qwen_conditional}
\end{table}
Larger models (Qwen3-VL-235B-A22B, Qwen-VL-Max) show consistent gains even when controlling for reconstruction quality, with Qwen-VL-Max approaching ceiling performance. The smaller Qwen3-VL-30B-A3B model does not benefit from tools in this setting, suggesting that effective tool orchestration requires sufficient model capacity.

Table~\ref{tab:qwen_family} reveals which question types benefit most from IVG across model scales.

\begin{table}[h]
\centering
\small
\begin{tabular}{lccc}
\toprule
Model & $\Delta$Agg & $\Delta$Comp & $\Delta$Topo \\
\midrule
Qwen3-VL-30B-A3B & $-0.002$ & $-0.094$ & $-0.032$ \\
Qwen3-VL-32B & $-0.003$ & $-0.001$ & $-0.074$ \\
Qwen3-VL-235B-A22B & $+0.001$ & $+0.007$ & $+0.008$ \\
Qwen-VL-Max & $0.000$ & $+0.003$ & $+0.008$ \\
\bottomrule
\end{tabular}
\caption{Conditional QA delta by question family ($S_{Data} \ge 0.9$). Agg=Aggregation, Comp=Comparison, Topo=Topology. $\Delta$ denotes Full $-$ Vision.}
\label{tab:qwen_family}
\end{table}

Capable models (Qwen3-VL-235B-A22B, Qwen-VL-Max) show their largest gains on Topology questions (+0.008), where view-grounded interaction (zoom, selection) helps resolve spatial relationships. This aligns with IVG's design goal of providing focal context for visually ambiguous regions. Effective use of IVG requires models with sufficient reasoning capacity to orchestrate tool calls alongside their primary tasks; in our evaluation, models with approximately 20B or more active parameters benefit consistently. For frontier models approaching ceiling performance, IVG's primary value shifts from accuracy improvement to auditability, providing explicit and traceable evidence for each reasoning step.
\section{IVG-Enabled Agents}
\label{sec:deepplot}

Beyond benchmark evaluation, we demonstrate that IVG enables a new class of visualization agents that can collaborate with users, explore data autonomously, and ground complex decisions in verifiable evidence.

\paragraph{Real-time collaboration.} In real-time collaboration, IVG bridges vague human gestures and precise machine reasoning. When a user points at a chart and asks "Why did it drop here?" (Figure~\ref{fig:interactive_workflow}), the question is inherently ambiguous. But the user's interaction history provides the missing context: the system captures the user's selection as focal context, telling the agent exactly which region the user is referring to. The agent retrieves the relevant data and provides a grounded response, enabling natural dialogue without the friction of verbal description. Details are in Appendix~\ref{sec:appendix_realtime}.

\begin{figure*}[!htbp]
    \centering
    \includegraphics[width=\textwidth]{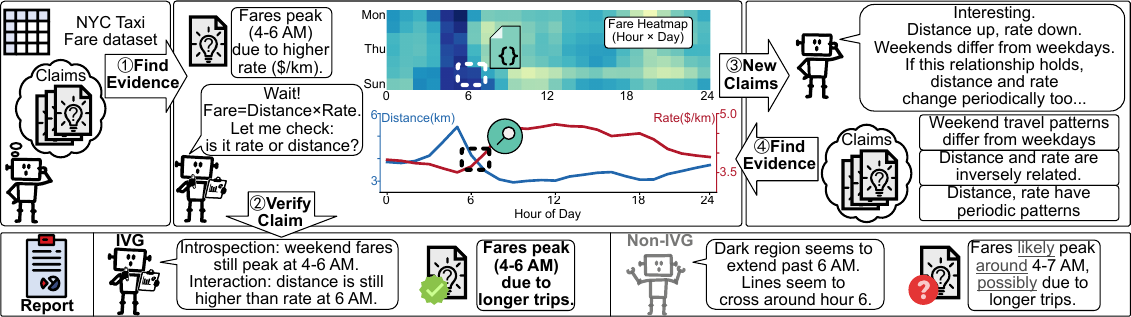}
    \caption{Autonomous exploration: IVG enables evidence-grounded analysis. The agent \ding{192}finds evidence for an initial claim, \ding{193}verifies it through IVG, discovers \ding{194}new claims, and \ding{195}gathers further evidence. Bottom: IVG yields a precise, verified claim (4-6 AM); without IVG, the agent produces a vague, incorrect one (4-7 AM).}
    \vspace{-5pt}
    \label{fig:autonomous_workflow}
\end{figure*}

\paragraph{Autonomous exploration with interactive deliverables.} 
In autonomous exploration, agents independently analyze datasets and produce evidence-grounded reports. The key challenge is verification: how can an agent ensure its claims are supported by data, rather than hallucinated?
Figure~\ref{fig:autonomous_workflow} illustrates the IVG cycle. Given a dataset, the agent forms initial claims, creates visualizations, and applies IVG to verify them: introspection retrieves exact values, while interaction focuses attention on relevant regions. Claims supported by deterministic evidence become grounded; the exploration process reveals new questions that feed back into the cycle. The final report contains only grounded claims, each traceable to visualization evidence. Crucially, the report remains interactive: users can ask follow-up questions grounded in their current view. Details and prompts are in Appendix~\ref{sec:appendix_autonomous}.

\paragraph{ML solution search.} 
\label{sec:ml_search}
The same IVG mechanisms extend from reporting to decision-making. In machine learning workflows, agents search over candidate solutions, drafting approaches, improving promising ones, and abandoning failures. Each candidate produces visualizations: training curves, confusion matrices, performance comparisons. The challenge is grounding search decisions in this accumulating evidence. IVG enables precise comparison across branches: rather than estimating from rendered images or relying on memory, the agent retrieves exact values from each candidate's visualization state. This transforms search decisions from heuristic judgments into evidence-grounded choices: which approaches to abandon, which to improve, and why.

\begin{figure*}[t]
    \centering
    \includegraphics[width=\textwidth]{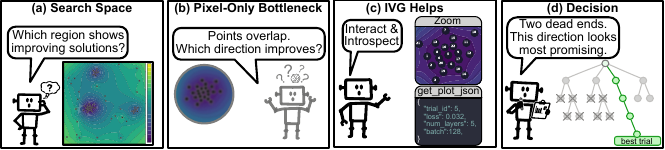}
    \caption{ML solution search. (a)~The agent surveys candidate solutions across a search space, but (b)~overlapping points make individual trials indistinguishable from pixels alone. (c)~View-grounded interaction zooms into a promising region and spec-grounded introspection retrieves exact trial metrics. (d)~IVG guides the agent to prune dead ends and focus on the most promising direction.}
    \label{fig:ml_search_appendix}
\end{figure*}

Figure~\ref{fig:ml_search_appendix} illustrates this process. The agent surveys candidate solutions across a search space, but overlapping points make individual trials indistinguishable from pixels alone. View-grounded interaction zooms into a promising region and spec-grounded introspection retrieves exact trial metrics, guiding the agent to prune dead ends and focus on the most promising direction.

The ML solution search uses Monte Carlo Tree Search (MCTS) to explore candidate solutions. Each node in the search tree represents a candidate solution, storing: a git commit hash (capturing the full code state), evaluation results (validation metrics), the agent's plan, and the run command used for execution. Multiple workers operate in parallel using git worktrees, which provide isolated working directories branched from the same repository. Each worker independently selects a node to expand (using UCT-based selection), checks out the parent's code state, generates improvements, executes the solution, and commits the new state. This parallelization accelerates search while git ensures reproducibility. Two agent types operate within the search tree: draft agents create new solutions from scratch (expanding from the root), and improve agents refine existing working solutions. Both agents have access to IVG tools, enabling them to create visualizations during development and retrieve exact values when comparing against previous attempts. Implementation details including prompt templates and the search interface are in Appendix~\ref{sec:appendix_ml_search}.

\paragraph{User study. }
To evaluate whether IVG improves agent-assisted analysis in practice, we conducted a blinded within-subjects study with 12 researchers with data analysis experience. Each participant used both an IVG-enabled agent and a vision-only baseline (in randomized order) to explore the same dataset, then completed a survey comparing the two.
11 of 12 participants preferred the IVG-enabled agent for interactive data exploration. Participants rated the preferred agent highly on citing correct values (4.5/5) and grounding claims in visible evidence (4.4/5). When compared against alternatives including general-purpose chatbots and manual analysis, 9 participants rated their preferred agent as the best option. Full survey design and per-participant results are in Appendix~\ref{sec:appendix_userstudy}.    

\section{Discussion}
Our benchmark results demonstrate that IVG addresses the Pixel-Only Bottleneck, and Section~\ref{sec:deepplot} shows these benefits extend to diverse real-world scenarios. 

\paragraph{Tool synergy and focal context.}  
A critical insight from our study is the synergy between spec-grounded introspection and view-grounded interaction. Introspection provides exact values but can overwhelm an agent with irrelevant structural details. Interaction provides focal context that localizes which subset of the specification is relevant, mimicking how a human analyst focuses attention before inspecting values. As shown in Section~\ref{sec:usage_analysis}, agents learn this balance: when introspection is available, total interaction calls drop by 71\% (Table~\ref{tab:tool_usage}), as agents substitute spec queries for view manipulation. However, when agents fail to make this substitution, unnecessary interactions consume the tool-call budget without providing new information, explaining why Full occasionally underperforms \textit{+Intro} on $S_{Data}$ (Table~\ref{tab:task1}).

\paragraph{Generality beyond plotly.} 
\label{sec:generality}
IVG builds on a property shared across visualization ecosystems: the separation of specification from rendering. Any library that maintains a structured specification can support spec-grounded introspection; any library that exposes interaction events as structured state can support view-grounded interaction. Plotly provides complete support for both, making it a natural first target. Other libraries offer partial coverage: Vega-Lite and ECharts maintain structured specifications with interaction events, while Matplotlib supports specification access but lacks native interaction. As visualization agents expand to these and other libraries, the same two-mechanism decomposition can guide tool design, identifying what specifications agents can read and what interactions produce structured context.

\section{Conclusion}
We present Introspective and Interactive Visual Grounding (IVG), a framework that gives visualization agents structured access to their own outputs. Evaluation on iPlotBench shows that introspection and interaction are complementary: one provides deterministic evidence, the other provides focal context. Tool usage analysis confirms that agents balance these mechanisms based on chart complexity. Beyond the benchmark, IVG enables agents to collaborate with users, explore data autonomously, and ground decisions in verifiable evidence.

\section{Limitations}
Our empirical validation uses Plotly as the target library. While the underlying abstraction generalizes to other grammar-based libraries (Section~\ref{sec:generality}), adapting the Visualization State API to each library's specification format and event model requires engineering effort. Our benchmark evaluates IVG through binary QA, which enables fully deterministic evaluation but does not capture the full range of analytical tasks. Section~\ref{sec:deepplot} demonstrates IVG on open-ended tasks qualitatively; a systematic open-ended benchmark is future work.

\bibliography{custom}

@inproceedings{yangchartmimic,
  title={ChartMimic: Evaluating LMM's Cross-Modal Reasoning Capability via Chart-to-Code Generation},
  author={Yang, Cheng and Shi, Chufan and Liu, Yaxin and Shui, Bo and Wang, Junjie and Jing, Mohan and XU, Linran and Zhu, Xinyu and Li, Siheng and Zhang, Yuxiang and others},
  booktitle={The Thirteenth International Conference on Learning Representations},
  year={2025},
}

@inproceedings{figureqa,
  title={FigureQA: An Annotated Figure Dataset for Visual Reasoning},
  author={Kahou, Samira Ebrahimi and Atkinson, Adam and Michalski, Vincent and K{\'a}d{\'a}r, {\'A}kos and Trischler, Adam and Bengio, Yoshua},
  booktitle={International Conference on Learning Representations (ICLR) Workshop},
  year={2018}
}

@inproceedings{masry2022chartqa,
  title={Chartqa: A benchmark for question answering about charts with visual and logical reasoning},
  author={Masry, Ahmed and Do, Xuan Long and Tan, Jia Qing and Joty, Shafiq and Hoque, Enamul},
  booktitle={Findings of the association for computational linguistics: ACL 2022},
  pages={2263--2279},
  year={2022}
}

@inproceedings{methani2020plotqa,
  title={Plotqa: Reasoning over scientific plots},
  author={Methani, Nitesh and Ganguly, Pritha and Khapra, Mitesh M and Kumar, Pratyush},
  booktitle={Proceedings of the ieee/cvf winter conference on applications of computer vision},
  pages={1527--1536},
  year={2020}
}

@inproceedings{kafle2018dvqa,
  title={Dvqa: Understanding data visualizations via question answering},
  author={Kafle, Kushal and Price, Brian and Cohen, Scott and Kanan, Christopher},
  booktitle={Proceedings of the IEEE conference on computer vision and pattern recognition},
  pages={5648--5656},
  year={2018}
}

@inproceedings{wu2025plot2code,
  title={Plot2code: A comprehensive benchmark for evaluating multi-modal large language models in code generation from scientific plots},
  author={Wu, Chengyue and Liang, Zhixuan and Ge, Yixiao and Guo, Qiushan and Lu, Zeyu and Wang, Jiahao and Shan, Ying and Luo, Ping},
  booktitle={Findings of the Association for Computational Linguistics: NAACL 2025},
  pages={3006--3028},
  year={2025}
}

@article{zhao2025chartcoder,
  title={Chartcoder: Advancing multimodal large language model for chart-to-code generation},
  author={Zhao, Xuanle and Luo, Xianzhen and Shi, Qi and Chen, Chi and Wang, Shuo and Liu, Zhiyuan and Sun, Maosong},
  journal={arXiv preprint arXiv:2501.06598},
  year={2025}
}

@inproceedings{zhang2024codeagent,
  title={CodeAgent: Enhancing Code Generation with Tool-Integrated Agent Systems for Real-World Repo-level Coding Challenges},
  author={Zhang, Kechi and Li, Jia and Li, Ge and Shi, Xianjie and Jin, Zhi},
  booktitle={Proceedings of the 62nd Annual Meeting of the Association for Computational Linguistics (Volume 1: Long Papers)},
  pages={13643--13658},
  year={2024}
}

@inproceedings{yang2024sweagent,
  title={{SWE}-agent: Agent-Computer Interfaces Enable Automated Software Engineering},
  author={John Yang and Carlos E Jimenez and Alexander Wettig and Kilian Lieret and Shunyu Yao and Karthik R Narasimhan and Ofir Press},
  booktitle={The Thirty-eighth Annual Conference on Neural Information Processing Systems},
  year={2024},
  url={https://arxiv.org/abs/2405.15793}
}

@InProceedings{guorepoaudit,
  title = 	 {{R}epo{A}udit: An Autonomous {LLM}-Agent for Repository-Level Code Auditing},
  author =       {Guo, Jinyao and Wang, Chengpeng and Xu, Xiangzhe and Su, Zian and Zhang, Xiangyu},
  booktitle = 	 {Proceedings of the 42nd International Conference on Machine Learning},
  pages = 	 {21083--21100},
  year = 	 {2025},
  editor = 	 {Singh, Aarti and Fazel, Maryam and Hsu, Daniel and Lacoste-Julien, Simon and Berkenkamp, Felix and Maharaj, Tegan and Wagstaff, Kiri and Zhu, Jerry},
  volume = 	 {267},
  series = 	 {Proceedings of Machine Learning Research},
  month = 	 {13--19 Jul},
  publisher =    {PMLR},
  url = 	 {https://proceedings.mlr.press/v267/guo25n.html}
}

@article{wang2024opendevin,
  title={Openhands: An open platform for ai software developers as generalist agents},
  author={Wang, Xingyao and Li, Boxuan and Song, Yufan and Xu, Frank F and Tang, Xiangru and Zhuge, Mingchen and Pan, Jiayi and Song, Yueqi and Li, Bowen and Singh, Jaskirat and others},
  journal={arXiv preprint arXiv:2407.16741},
  year={2024}
}

@inproceedings{zhang2024autocoderover,
  title={Autocoderover: Autonomous program improvement},
  author={Zhang, Yuntong and Ruan, Haifeng and Fan, Zhiyu and Roychoudhury, Abhik},
  booktitle={Proceedings of the 33rd ACM SIGSOFT International Symposium on Software Testing and Analysis},
  pages={1592--1604},
  year={2024}
}

@article{xia2024agentless,
  title={Agentless: Demystifying LLM-based Software Engineering Agents},
  author={Xia, Chunqiu Steven and Deng, Yinlin and Dunn, Soren and Zhang, Lingming},
  journal={CoRR},
  year={2024}
}

@inproceedings{niu2025chart2code53,
  title={Chart2Code53: A Large-Scale Diverse and Complex Dataset for Enhancing Chart-to-Code Generation},
  author={Niu, Tianhao and Cui, Yiming and Wang, Baoxin and Xu, Xiao and Yao, Xin and Zhu, Qingfu and Wu, Dayong and Wang, Shijin and Che, Wanxiang},
  booktitle={Proceedings of the 2025 Conference on Empirical Methods in Natural Language Processing},
  pages={15839--15855},
  year={2025}
}

@misc{seo2025automatedvisualizationcodesynthesis,
      title={Automated Visualization Code Synthesis via Multi-Path Reasoning and Feedback-Driven Optimization}, 
      author={Wonduk Seo and Seungyong Lee and Daye Kang and Hyunjin An and Zonghao Yuan and Seunghyun Lee},
      year={2025},
      eprint={2502.11140},
      archivePrefix={arXiv},
      primaryClass={cs.SE},
      url={https://arxiv.org/abs/2502.11140}, 
}

@inproceedings{hong-etal-2025-data,
    title = "Data Interpreter: An {LLM} Agent for Data Science",
    author = "Hong, Sirui  and
      Lin, Yizhang  and
      Liu, Bang  and
      Liu, Bangbang  and
      Wu, Binhao  and
      Zhang, Ceyao  and
      Li, Danyang  and
      Chen, Jiaqi  and
      Zhang, Jiayi  and
      Wang, Jinlin  and
      Zhang, Li  and
      Zhang, Lingyao  and
      Yang, Min  and
      Zhuge, Mingchen  and
      Guo, Taicheng  and
      Zhou, Tuo  and
      Tao, Wei  and
      Tang, Robert  and
      Lu, Xiangtao  and
      Zheng, Xiawu  and
      Liang, Xinbing  and
      Fei, Yaying  and
      Cheng, Yuheng  and
      Ni, Yongxin  and
      Gou, Zhibin  and
      Xu, Zongze  and
      Luo, Yuyu  and
      Wu, Chenglin",
    editor = "Che, Wanxiang  and
      Nabende, Joyce  and
      Shutova, Ekaterina  and
      Pilehvar, Mohammad Taher",
    booktitle = "Findings of the Association for Computational Linguistics: ACL 2025",
    month = jul,
    year = "2025",
    address = "Vienna, Austria",
    publisher = "Association for Computational Linguistics",
    url = "https://aclanthology.org/2025.findings-acl.1016/",
    doi = "10.18653/v1/2025.findings-acl.1016",
    pages = "19796--19821",
    ISBN = "979-8-89176-256-5"
}

@inproceedings{kaul2024throne,
  title={Throne: An object-based hallucination benchmark for the free-form generations of large vision-language models},
  author={Kaul, Prannay and Li, Zhizhong and Yang, Hao and Dukler, Yonatan and Swaminathan, Ashwin and Taylor, CJ and Soatto, Stefano},
  booktitle={Proceedings of the IEEE/CVF Conference on Computer Vision and Pattern Recognition},
  pages={27228--27238},
  year={2024}
}

@inproceedings{jiang2024hallucination,
  title={Hallucination augmented contrastive learning for multimodal large language model},
  author={Jiang, Chaoya and Xu, Haiyang and Dong, Mengfan and Chen, Jiaxing and Ye, Wei and Yan, Ming and Ye, Qinghao and Zhang, Ji and Huang, Fei and Zhang, Shikun},
  booktitle={Proceedings of the IEEE/CVF Conference on Computer Vision and Pattern Recognition},
  pages={27036--27046},
  year={2024}
}

@inproceedings{rohrbach2018object,
  title={Object Hallucination in Image Captioning},
  author={Rohrbach, Anna and Hendricks, Lisa Anne and Burns, Kaylee and Darrell, Trevor and Saenko, Kate},
  booktitle={Proceedings of the 2018 Conference on Empirical Methods in Natural Language Processing},
  pages={4035--4045},
  year={2018}
}

@article{li2025metal,
  title={METAL: A Multi-Agent Framework for Chart Generation with Test-Time Scaling},
  author={Li, Bingxuan and Wang, Yiwei and Gu, Jiuxiang and Chang, Kai-Wei and Peng, Nanyun},
  journal={CoRR},
  year={2025}
}

@article{schick2023toolformer,
  title={Toolformer: Language models can teach themselves to use tools},
  author={Schick, Timo and Dwivedi-Yu, Jane and Dess{\`\i}, Roberto and Raileanu, Roberta and Lomeli, Maria and Hambro, Eric and Zettlemoyer, Luke and Cancedda, Nicola and Scialom, Thomas},
  journal={Advances in Neural Information Processing Systems},
  volume={36},
  pages={68539--68551},
  year={2023}
}

@inproceedings{yao2022react,
  title={React: Synergizing reasoning and acting in language models},
  author={Yao, Shunyu and Zhao, Jeffrey and Yu, Dian and Du, Nan and Shafran, Izhak and Narasimhan, Karthik R and Cao, Yuan},
  booktitle={The eleventh international conference on learning representations},
  year={2022}
}

@inproceedings{li2023evaluating,
  title={Evaluating Object Hallucination in Large Vision-Language Models},
  author={Li, Yifan and Du, Yifan and Zhou, Kun and Wang, Jinpeng and Zhao, Wayne Xin and Wen, Ji-Rong},
  booktitle={Proceedings of the 2023 Conference on Empirical Methods in Natural Language Processing},
  pages={292--305},
  year={2023}
}

@inproceedings{leng2024mitigating,
  title={Mitigating object hallucinations in large vision-language models through visual contrastive decoding},
  author={Leng, Sicong and Zhang, Hang and Chen, Guanzheng and Li, Xin and Lu, Shijian and Miao, Chunyan and Bing, Lidong},
  booktitle={Proceedings of the IEEE/CVF Conference on Computer Vision and Pattern Recognition},
  pages={13872--13882},
  year={2024}
}

@article{satyanarayan2016vega,
  title={Vega-lite: A grammar of interactive graphics},
  author={Satyanarayan, Arvind and Moritz, Dominik and Wongsuphasawat, Kanit and Heer, Jeffrey},
  journal={IEEE transactions on visualization and computer graphics},
  volume={23},
  number={1},
  pages={341--350},
  year={2016},
  publisher={IEEE}
}

@book{zhu2013data,
  title={Data visualization with D3. js cookbook},
  author={Zhu, Nick Qi},
  year={2013},
  publisher={Packt Publishing Ltd}
}

@article{wickham2011ggplot2,
  title={ggplot2},
  author={Wickham, Hadley},
  journal={Wiley interdisciplinary reviews: computational statistics},
  volume={3},
  number={2},
  pages={180--185},
  year={2011},
  publisher={Wiley Online Library}
}

@incollection{wilkinson2011grammar,
  title={The grammar of graphics},
  author={Wilkinson, Leland},
  booktitle={Handbook of computational statistics: Concepts and methods},
  pages={375--414},
  year={2011},
  publisher={Springer}
}

@ARTICLE{Voyager1,
  author={Wongsuphasawat, Kanit and Moritz, Dominik and Anand, Anushka and Mackinlay, Jock and Howe, Bill and Heer, Jeffrey},
  journal={IEEE Transactions on Visualization and Computer Graphics}, 
  title={Voyager: Exploratory Analysis via Faceted Browsing of Visualization Recommendations}, 
  year={2016},
  volume={22},
  number={1},
  pages={649-658},
  doi={10.1109/TVCG.2015.2467191}}

@inproceedings{Voyager2,
author = {Wongsuphasawat, Kanit and Qu, Zening and Moritz, Dominik and Chang, Riley and Ouk, Felix and Anand, Anushka and Mackinlay, Jock and Howe, Bill and Heer, Jeffrey},
title = {Voyager 2: Augmenting Visual Analysis with Partial View Specifications},
year = {2017},
isbn = {9781450346559},
publisher = {Association for Computing Machinery},
address = {New York, NY, USA},
url = {https://doi.org/10.1145/3025453.3025768},
doi = {10.1145/3025453.3025768},
booktitle = {Proceedings of the 2017 CHI Conference on Human Factors in Computing Systems},
pages = {2648–2659},
numpages = {12},
location = {Denver, Colorado, USA},
series = {CHI '17}
}

@INPROCEEDINGS{draco,
  author={Yang, Junran and Gyarmati, Péter Ferenc and Zeng, Zehua and Moritz, Dominik},
  booktitle={2023 IEEE Visualization and Visual Analytics (VIS)}, 
  title={Draco 2: An Extensible Platform to Model Visualization Design}, 
  year={2023},
  volume={},
  number={},
  pages={166-170},
  doi={10.1109/VIS54172.2023.00042}}

@ARTICLE{6634168,
  author={Brehmer, Matthew and Munzner, Tamara},
  journal={IEEE Transactions on Visualization and Computer Graphics}, 
  title={A Multi-Level Typology of Abstract Visualization Tasks}, 
  year={2013},
  volume={19},
  number={12},
  pages={2376-2385},
  doi={10.1109/TVCG.2013.124}}

\appendix
\section{Visualization State API: MCP Tool Schemas}
\label{sec:appendix_api}

The Visualization State API is implemented as a set of Model Context Protocol (MCP) tools. MCP provides a standardized interface for language model agents to invoke external functions with typed arguments and structured JSON responses. All tools return JSON-formatted output for deterministic parsing.

\subsection{Base Tools}

\paragraph{show\_plot}
Creates a Plotly figure from agent-generated Python code and registers it for subsequent operations.
\begin{itemize}
    \item \textbf{Arguments:}
    \begin{itemize}
        \item plotly\_codes (string, required): Python code defining a variable fig as a Plotly figure object. Must include necessary imports.
    \end{itemize}
    \item \textbf{Returns:} A unique identifier for referencing this figure in subsequent tool calls.
    \begin{quote}
    \small \{plot\_id: int\}
    \end{quote}
\end{itemize}

\paragraph{get\_plot\_image}
Captures the current rendered view of a figure as a PNG screenshot.
\begin{itemize}
    \item \textbf{Arguments:}
    \begin{itemize}
        \item plot\_id (integer, required): Plot identifier from show\_plot.
        \item interaction\_id (integer, optional): Specific interaction snapshot ID. If omitted, returns the latest view state.
    \end{itemize}
    \item \textbf{Returns:} Absolute path to the PNG file, which can be passed to the VLM for pixel-based inspection.
    \begin{quote}
    \small \{image\_path: str\}
    \end{quote}
\end{itemize}

\subsection{Spec-Grounded Introspection}

\paragraph{get\_plot\_json}
Retrieves the full Plotly specification of the agent's constructed figure, enabling deterministic verification of data values and visual encodings.
\begin{itemize}
    \item \textbf{Arguments:}
    \begin{itemize}
        \item plot\_id (integer, required): Plot identifier from show\_plot.
    \end{itemize}
    \item \textbf{Returns:} The complete Plotly JSON specification, where data contains trace objects and layout contains axis and styling configurations.
    \begin{quote}
    \small \{data: [...], layout: \{...\}\}
    \end{quote}
\end{itemize}

\noindent\textbf{Trust Boundary:} This tool exposes \emph{only} the agent's own constructed figure specification---not the ground-truth dataset or reference solution.

\subsection{View-Grounded Interaction}

\paragraph{relayout}
Programmatically zooms or pans the view by setting axis ranges, providing focal context for dense or occluded regions.
\begin{itemize}
    \item \textbf{Arguments:}
    \begin{itemize}
        \item plot\_id (integer, required): Plot identifier.
        \item x\_min, x\_max (number, optional): X-axis range bounds.
        \item y\_min, y\_max (number, optional): Y-axis range bounds.
    \end{itemize}
    \item \textbf{Returns:} Success status.
    \item \textbf{Event Payload:} Records axis ranges.
    \begin{quote}
    \small
    \{\\
    \hspace*{1em}``xaxis.range[0]'': 1981.97,\\
    \hspace*{1em}``xaxis.range[1]'': 2001.99,\\
    \hspace*{1em}``yaxis.range[0]'': 70.98,\\
    \hspace*{1em}``yaxis.range[1]'': 73.81\\
    \}
    \end{quote}
\end{itemize}

\paragraph{legendclick}
Toggles the visibility of a specific trace, enabling isolation of overlapping series.
\begin{itemize}
    \item \textbf{Arguments:}
    \begin{itemize}
        \item plot\_id (integer, required): Plot identifier.
        \item curve\_number (integer, required): Zero-indexed trace index.
    \end{itemize}
    \item \textbf{Returns:} Success status.
    \item \textbf{Event Payload:} Records the curve number. Visibility changes are captured in the associated plotly\_restyle event.
    \begin{quote}
    \small
    \{\\
    \hspace*{1em}curve\_number: int,\\
    \hspace*{1em}expanded\_index: int\\
    \}
    \end{quote}
\end{itemize}

\paragraph{selected}
Performs a box selection over a specified region and returns information about selected data points.
\begin{itemize}
    \item \textbf{Arguments:}
    \begin{itemize}
        \item plot\_id (integer, required): Plot identifier.
        \item x\_min, x\_max (number, optional): X-axis selection bounds.
        \item y\_min, y\_max (number, optional): Y-axis selection bounds.
    \end{itemize}
    \item \textbf{Returns:} Selection information.
    \item \textbf{Event Payload:} Records the selection range and point count.
    \begin{quote}
    \small
    \{\\
    \hspace*{1em}point\_count: int,\\
    \hspace*{1em}range: \{\\
    \hspace*{2em}x: [min, max],\\
    \hspace*{2em}y: [min, max]\\
    \hspace*{1em}\}\\
    \}
    \end{quote}
\end{itemize}

\paragraph{query\_interactions}
Retrieves the interaction history for a figure, enabling the agent to reason about prior view manipulations.
\begin{itemize}
    \item \textbf{Arguments:}
    \begin{itemize}
        \item plot\_id (integer, required): Plot identifier.
        \item event\_type (string, optional): Filter by event type (init, relayout, legendclick, selected).
    \end{itemize}
    \item \textbf{Returns:} Chronological list of interaction events, each with:
    \begin{itemize}
        \item id: Unique interaction ID (can be passed to get\_plot\_image to retrieve historical snapshots).
        \item event\_type: One of init, relayout, legendclick, selected.
        \item payload: Event-specific details (e.g., axis ranges for relayout, visibility state for legendclick).
        \item has\_screenshot: Whether a screenshot was captured for this interaction state.
    \end{itemize}
    \begin{quote}
    \small
    \{\\
    \hspace*{1em}events: [\\
    \hspace*{2em}\{id, event\_type, payload, has\_screenshot\},\\
    \hspace*{2em}...\\
    \hspace*{1em}]\\
    \}
    \end{quote}
\end{itemize}

\section{Controller and Evaluation Protocol}
\label{sec:appendix_controller}

This section details the prompt templates, stop conditions, and execution environment used in our evaluation.

\subsection{Prompt Templates}

All agent configurations share identical prompt templates. We use minimal prompts to avoid biasing tool usage patterns.

\paragraph{Task 1: Chart Recreation}
The agent receives the reference image (base64-encoded PNG) along with the following text prompt:

\begin{quote}
\small
Read ./input.png and recreate this plot.

Output the Plotly figure as JSON with ``data'' and ``layout'' keys: \{``data'': [...], ``layout'': \{...\}\}
\end{quote}

\noindent For configurations with tools enabled, the agent may invoke Visualization State API tools before producing the final JSON output. For the Vision baseline, the agent directly outputs JSON without tool access.

\paragraph{Task 2: Visual QA}
For each binary question, the agent receives:

\begin{quote}
\small
<question text>

Reply with ONLY a single digit: 0 or 1
\end{quote}

\noindent Task 2 questions are answered within the same session as Task 1, meaning the agent retains context from chart recreation and can reference its constructed figure state.

\subsection{Stop Conditions}

Each \textit{task invocation} is bounded by two termination conditions:

\begin{itemize}
    \item \textbf{Tool round limit:} Maximum 5 tool-calling rounds per invocation. If the model continues requesting tool calls after 5 rounds, the invocation terminates with partial results.
    \item \textbf{API timeout:} 120 seconds per API call. If the model does not respond within this window, the invocation is marked as failed.
\end{itemize}

\noindent\textbf{Invocation structure:} An episode consists of multiple invocations within a persistent session. Task~1 (chart recreation) is a single invocation. Task~2 questions are each processed as \textit{separate invocations}, meaning each question receives its own 5-round tool budget. Critically, the session context (conversation history and figure state) persists across all invocations, allowing the agent to reference its Task~1 reconstruction when answering Task~2 questions.

\noindent An invocation terminates successfully when at least one valid answer is extracted: a Plotly JSON object with non-empty data array for Task~1, or a parseable 0/1 response for Task~2.

\subsection{Execution Environment}

\paragraph{Claude Agents (Main Ablation).}
Claude 4.5 Haiku agents run in isolated Docker containers based on ``python:3.10-slim''. Each container includes:
\begin{itemize}
    \item Python 3.10 with Plotly for figure generation
    \item Kaleido for server-side PNG rendering
    \item MCP server implementing the Visualization State API
    \item Isolated filesystem with read-only access to reference images
\end{itemize}

\noindent The containerized environment ensures reproducibility and prevents agents from accessing ground-truth data or other test cases.

\paragraph{Qwen-VL Agents (Scaling Study).}
Qwen-VL models are accessed via OpenAI-compatible APIs (local vLLM or OpenRouter). The same MCP tools are available, with images passed as base64-encoded PNGs in the image\_url message format. Key differences from the Claude setup:
\begin{itemize}
    \item No Docker isolation (API-based execution)
    \item Session state managed by Python runner script
    \item Same prompt templates and tool schemas
    \item Vendor runtimes may include undisclosed system prompts
\end{itemize}

\noindent To account for potential confounds from vendor-specific behaviors, we emphasize within-model comparisons (Vision vs Full) rather than cross-model rankings.

\subsection{Tool Schema Availability}

In all tool-enabled configurations (+Inter, +Intro, Full), the complete tool schema is provided to the model at session start. Tools are \textit{available} but their use is \textit{optional}---models autonomously decide when and whether to invoke tools based on task requirements. This design isolates model capability from prompt engineering that might bias tool usage.

\section{Deployment Details}
\label{sec:appendix_deployments}

This section provides additional implementation details for the deployments described in Section~\ref{sec:deepplot}.

\begin{figure*}[htbp]
    \centering
    \includegraphics[width=\textwidth]{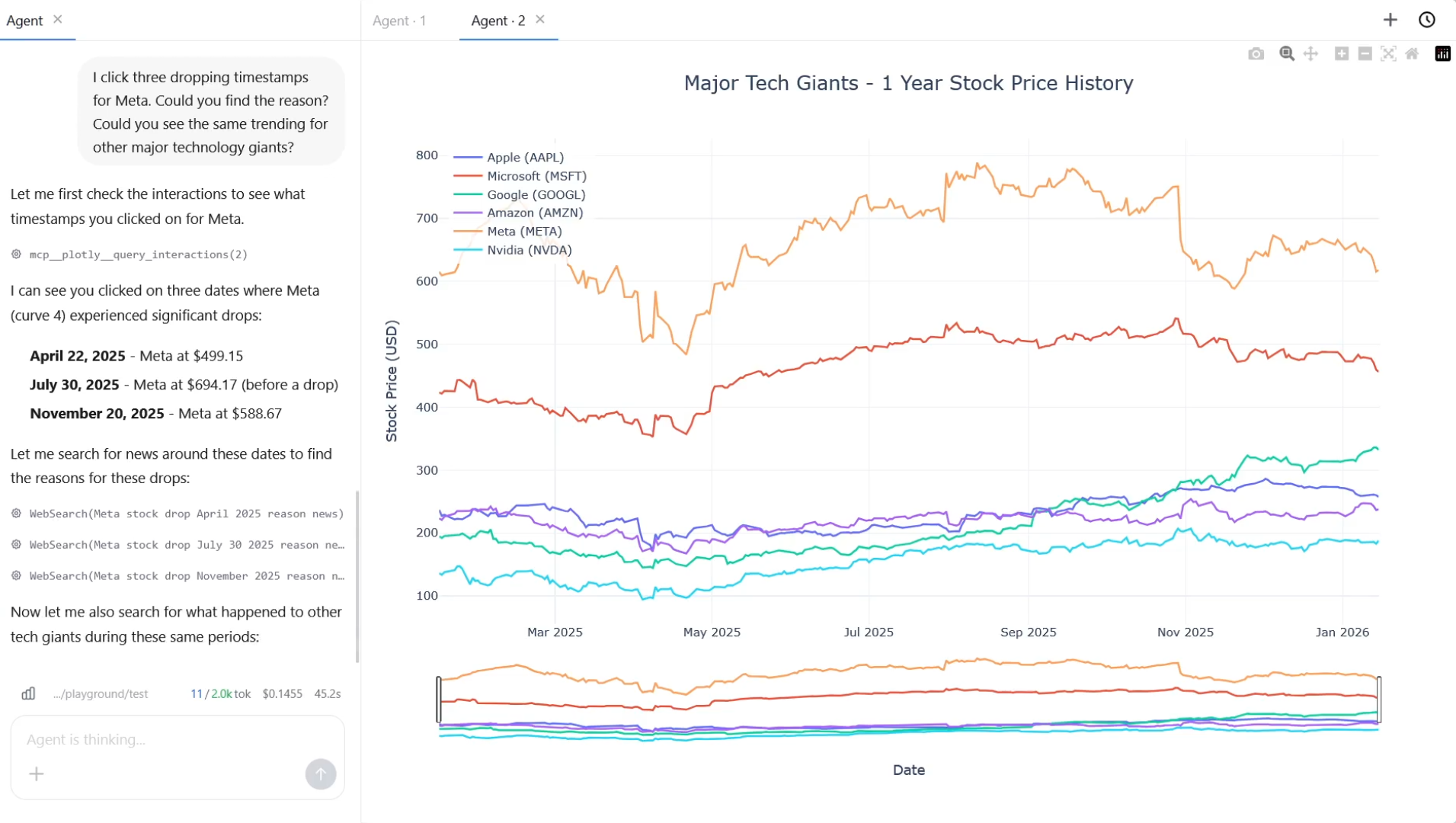}
    \caption{Real-time collaboration interface. Left: chat panel where users interact with the agent. Right: interactive Plotly visualizations created by the agent. Users can manipulate plots directly, and their interaction history informs subsequent agent responses.}
    \label{fig:screenshot_agent}
\end{figure*}

\subsection{Real-time Collaboration}

\label{sec:appendix_realtime}

In real-time collaboration mode, IVG bridges vague human gestures and precise machine reasoning. When a user points at a chart and asks ``What's happening here?'', the question is inherently ambiguous, but the user's interaction history provides the missing context. As users explore a visualization, view-grounded interaction captures each action as structured context that the agent can retrieve and interpret.

For example, a user viewing a stock market chart notices a sharp drop and asks ``Why did it drop here?'' Rather than verbally describing the region of interest, the user simply points and asks. The system captures the user's selection as focal context, telling the agent exactly which region the user is referring to. The agent can then retrieve the relevant data and provide a grounded response. This transforms the agent from a passive chart reader into an active collaborator that shares the user's visual attention, enabling natural dialogue without the friction of verbal description.

Figure~\ref{fig:screenshot_agent} illustrates a concrete example of IVG-enabled agent working in real-time collaboration. The user clicks three timestamps on a stock price chart where Meta experienced drops and asks the agent to investigate. The agent begins by calling \texttt{query\_interactions} to retrieve the user's click history, identifying the exact dates and prices: April 22 at \$499.15, July 30 at \$694.17, and November 20 at \$588.67. These values come directly from the chart specification, not from pixel estimation. The agent then uses web search to find explanations for each drop and extends the analysis to other tech giants during the same periods. This demonstrates how IVG bridges the user's visual gesture (clicking on drops) with precise, verifiable data retrieval, enabling a grounded analytical dialogue.

\subsection{Autonomous Exploration}
\label{sec:appendix_autonomous}

The autonomous exploration agent (Deep Plot) operates in a time-bounded loop, iteratively analyzing data until timeout. IVG serves a dual role in this cycle. First, introspection verifies whether evidence supports a claim: the agent retrieves exact values rather than estimating from rendered pixels. Second, interaction enables discovery: by manipulating views to focus on specific regions, the agent uncovers patterns that prompt new questions. The same mechanism that grounds existing claims also expands the space of inquiry.

Three prompts guide the process.

\paragraph{Initial Prompt.}
The agent begins with an exploration phase:
\begin{quote}
\small
Explore the working directory and find data files to analyze. You have \{timeout\_seconds\} seconds.
\textbf{Research process:}
\begin{enumerate}
    \item \textbf{Look} -- List files in the directory. Load any data files you find (csv, json, parquet, etc.).
    \item \textbf{Ask} -- What is this dataset about? What questions are worth investigating?
    \item \textbf{Investigate} -- Answer your questions. Support each finding with evidence (numbers or plotly visualizations).
    \item \textbf{Synthesize} -- What did you learn?
\end{enumerate}
\end{quote}

\paragraph{Improvement Prompt.}
The agent iteratively refines its analysis:
\begin{quote}
\small
You have \{time\_remaining\} seconds remaining.
\textbf{Reflect:}
\begin{itemize}
    \item Does your evidence support your findings?
    \item Dig deeper. Check specific values. Use get\_plot\_image to view plots, and relayout/legendclick/selected to interact.
    \item What new questions emerge?
\end{itemize}
Continue investigating.
\end{quote}

\paragraph{Final Report Prompt.}
After timeout, the agent produces a structured report:
\begin{quote}
\small
Write your \textbf{Final Report}.
\textbf{Structure:}
\begin{enumerate}
    \item \textbf{Data} -- Describe the dataset
    \item \textbf{Findings} -- List each finding with evidence (numbers, tables, or plot references)
\end{enumerate}
\textbf{To finish:}
\begin{enumerate}
    \item Redraw any unclear plots (fix overlapping, ensure readable).
    \item Decide which plots are evidence and their display order (e.g., plots 4, 7, 2).
    \item Write ``analysis.md'' referencing plots as Plot 1, Plot 2, Plot 3... (matching the order you chose).
    \item Call submit\_summary(evidence\_plots=[4, 7, 2]) with original IDs in that order.
\end{enumerate}
\end{quote}

\begin{figure*}[t]
    \centering
    \includegraphics[width=\textwidth]{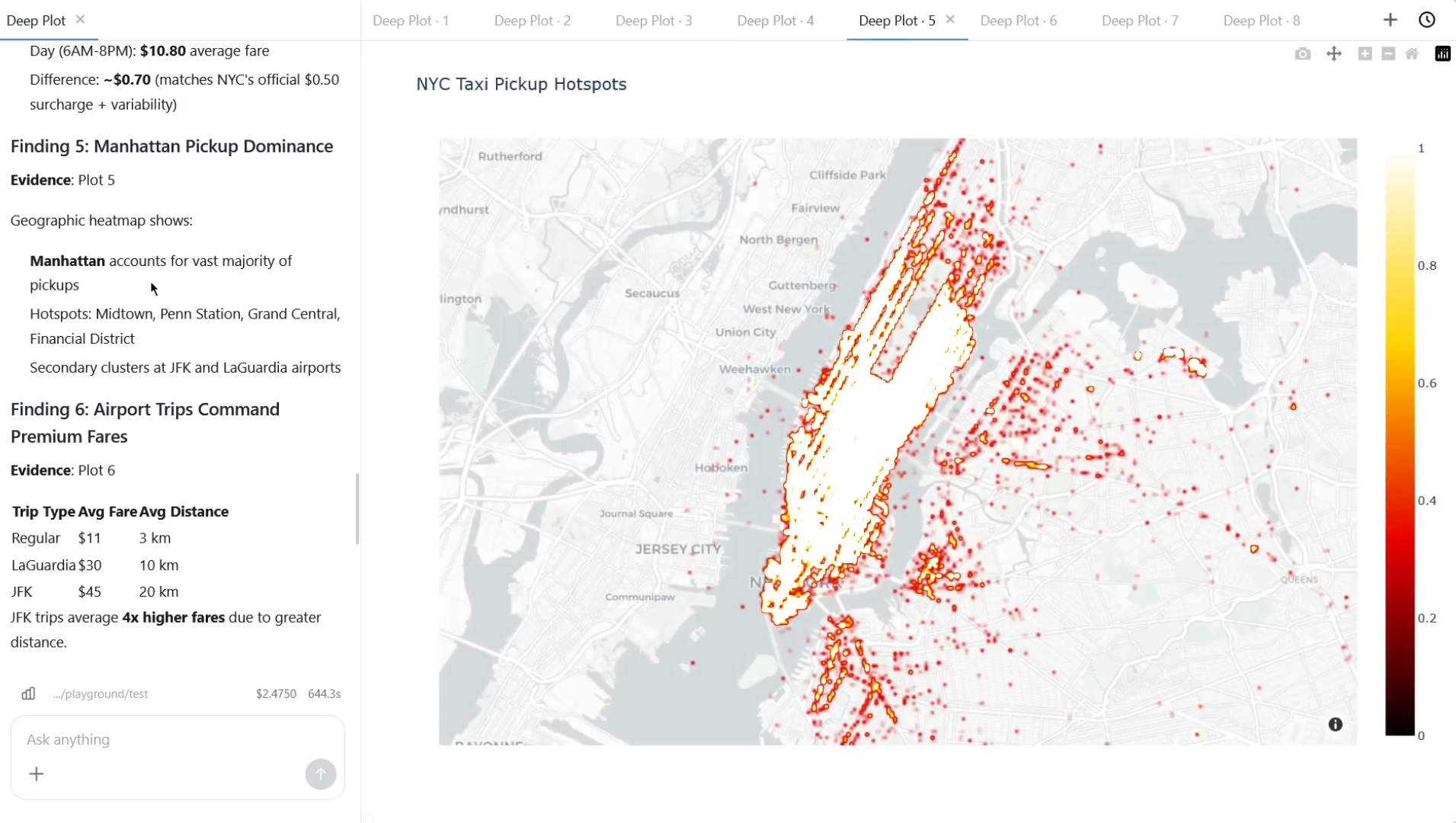}
    \caption{Autonomous exploration interface. Left: agent-generated analysis report with findings linked to supporting visualizations. Right: interactive Plotly figures in tabbed views. Users can interact with plots and ask follow-up questions after the autonomous analysis completes.}
    \label{fig:screenshot_deepplot}
\end{figure*}

\paragraph{Report Interface.}
Figure~\ref{fig:screenshot_deepplot} shows the final report interface with a dual-panel layout: a left panel containing the agent-generated report and a right panel hosting interactive Plotly figures in tabbed views. Each finding in the report links to a supporting plot, and the report contains only grounded claims, each traceable to visualization evidence. Crucially, the report remains interactive: users can interact with plots and ask follow-up questions grounded in their current view via the input field at the bottom.

For example, Finding~5 (``Manhattan Pickup Dominance'') is supported by a geographic heatmap (Plot~5, shown in the right panel), with the agent identifying specific hotspots: Midtown, Penn Station, Grand Central, and Financial District, plus secondary clusters at JFK and LaGuardia airports. Finding~6 (``Airport Trips Command Premium Fares'') cites a comparison table with exact values retrieved through spec-grounded introspection: regular trips average \$11 over 3~km, LaGuardia trips \$30 over 10~km, and JFK trips \$45 over 20~km, concluding that JFK fares are 4$\times$ higher due to greater distance.

\subsection{ML Solution Search}
\label{sec:appendix_ml_search}

\paragraph{Draft Prompt.}
Creates a new solution from scratch:
\begin{quote}
\small
You are solving an ML task.
\textbf{Task:} \{task\_description\}
\textbf{Data:} \{data\_report\}
\textbf{Output Requirements:} \{output\_requirements\}
\textbf{Goal:} \{goal\}
\textbf{Memory:} The memory of previous solutions used to solve this task is provided below: \{sibling\_memory\}
\textbf{Guidelines:}
\begin{itemize}
    \item When proposing your design, take the Memory section into account.
    \item Your proposed solution \textbf{must be distinctly different} from the existing designs in the Memory section.
    \item If a previous approach had bugs, try a different approach to avoid the same issues.
    \item Use MCP Plotly tools to interactively explore and visualize as needed to support your design.
\end{itemize}
\end{quote}

\paragraph{Improve Prompt.}
Refines an existing working solution:
\begin{quote}
\small
Improve this working ML solution.
\textbf{Task:} \{task\_description\}
\textbf{Current Solution (to improve):} \{parent\_memory\}
\textbf{Current Execution Output:} \{execution\_output\}
\textbf{Sibling Memory (Previous Improvement Attempts):} \{sibling\_memory\}
\textbf{Goal:} \{goal\}
\textbf{Guidelines:}
\begin{itemize}
    \item When proposing your improvement, take the Sibling Memory section into account.
    \item Your proposed improvement \textbf{must be distinctly different} from the existing attempts in the Sibling Memory section.
    \item Review the Current Execution Output to understand what worked and what could be improved.
    \item Use MCP Plotly tools to interactively explore and visualize as needed to support your improvement.
\end{itemize}
\end{quote}

\begin{figure*}[t]
    \centering
    \includegraphics[width=\textwidth]{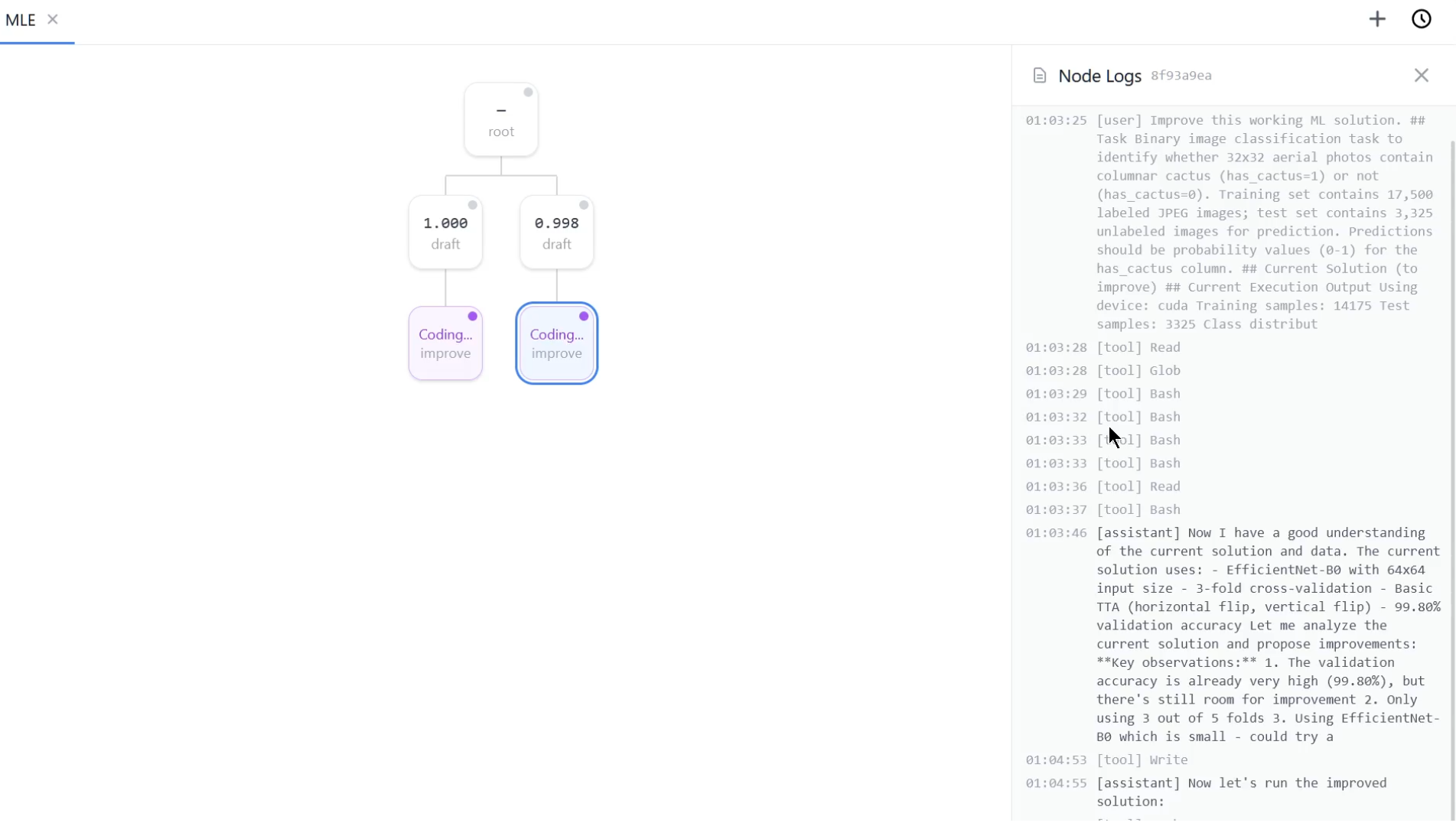}
    \caption{ML solution search interface. The tree visualization shows candidate solutions as nodes, with metrics displayed for each branch. Agents use IVG to create and compare training curves, confusion matrices, and other visualizations across branches, grounding search decisions in deterministic evidence.}
    \label{fig:screenshot_mle}
\end{figure*}

\paragraph{Search Interface.}
Figure~\ref{fig:screenshot_mle} shows the search tree interface for a binary image classification task (identifying columnar cactus in 32$\times$32 aerial photos). The left panel displays the search tree: a root node branches into two draft solutions (scoring 1.000 and 0.998), each of which spawns an improve node currently being refined. The right panel shows detailed logs for a selected node, revealing the agent's reasoning process. The agent first reads the current solution and data using tool calls (Read, Glob, Bash), then analyzes the approach: ``The current solution uses EfficientNet-B0 with 64$\times$64 input size, 3-fold cross-validation, basic TTA (horizontal flip, vertical flip), 99.80\% validation accuracy.'' Based on this analysis, the agent identifies specific improvement opportunities (increasing folds, trying a larger model) and writes an improved solution. Both draft and improve agents have access to IVG tools, enabling them to create visualizations during development and retrieve exact metrics when comparing against previous attempts.

\subsection{User Study}
\label{sec:appendix_userstudy}

\paragraph{Study Design.}
We conducted a blinded within-subjects study to evaluate IVG in practice. Each participant used two agents (IVG-enabled and vision-only baseline) to explore the same dataset. The agents were labeled Agent~A and Agent~B in randomized order to prevent bias. After using both agents, participants completed a survey comparing the two without knowing which agent used IVG.

\paragraph{Participants.}
We recruited 12 participants, all researchers with data analysis experience. All participants had prior experience with data visualization tools.

\paragraph{Task.}
Each participant was given a dataset of 3.86M NYC yellow taxi rides (January--June 2015) containing pickup/dropoff coordinates, timestamps, passenger counts, and fare amounts. Participants were asked to explore spatial, temporal, and pricing patterns through interactive charts, answer analytical questions about the data, and assess the quality of the agent's responses.

\paragraph{Survey Questions.}
After using both agents, participants answered the following:

\begin{itemize}
    \item \textbf{Q1 (Preference):} Which agent do you prefer for interactive data exploration? [Agent A / Agent B]
    \item \textbf{Q2--Q6 (Likert 1--5):} Rate your experience with your preferred agent:
    \begin{itemize}
        \item Q2: The agent correctly identified the region I interacted with.
        \item Q3: The agent cited correct numbers from the chart.
        \item Q4: Each claim in the report was supported by evidence visible in the charts.
        \item Q5: The agent found patterns I had not noticed before.
        \item Q6: The interactive charts helped me verify the agent's claims.
    \end{itemize}
    \item \textbf{Q7 (Comparison):} Compared to your preferred agent, would you rather use: [My preferred agent is the best option / General-purpose chatbot / Coding assistant / Manual analysis]
\end{itemize}

\paragraph{Results.}

\begin{table}[h]
\centering
\small
\begin{tabular}{lc}
\toprule
Question & Mean (SD) \\
\midrule
Q1: Preferred IVG agent & 11/12 participants \\
Q2: Correctly identified region & 4.9 (0.3) \\
Q3: Cited correct numbers & 4.5 (0.8) \\
Q4: Claims supported by evidence & 4.4 (0.7) \\
Q5: Found new patterns & 3.4 (1.2) \\
Q6: Interactive charts aided verification & 4.2 (0.8) \\
\midrule
Q7: Preferred agent is best overall option & 9/12 participants \\
\bottomrule
\end{tabular}
\caption{User study results. Likert ratings are on a 5-point scale (1 = Strongly Disagree, 5 = Strongly Agree).}
\label{tab:user_study_full}
\end{table}

Participants rated the IVG-enabled agent highest on region identification (4.9) and citing correct values (4.5), indicating that IVG's clearest benefit is precise grounding rather than open-ended discovery. The lower score on discovering new patterns (3.4) suggests that the system currently helps more with verification than with surfacing novel insights. The only participant who preferred Agent~B cited response style rather than factual quality, favoring shorter and easier-to-understand answers over detailed numeric explanations. The three participants who selected a general-purpose chatbot in Q7 did not reject IVG in the paired comparison; instead, they cited familiarity with existing canvas- or Matplotlib-based workflows and reported that interaction context was less central to their own analysis process.

\end{document}